\documentclass{article}
\usepackage{ijcai24}

\usepackage{times}
\usepackage{soul}
\usepackage{url}
\usepackage[hidelinks]{hyperref}
\usepackage[utf8]{inputenc}
\usepackage[small]{caption}
\usepackage{graphicx}
\usepackage{amsmath}
\usepackage{amsthm}
\usepackage{booktabs}
\usepackage{algorithm}
\usepackage{algorithmic}
\usepackage[switch]{lineno}

\usepackage{bbm}
\usepackage{amssymb}
\usepackage{amssymb}

\usepackage[table]{xcolor}
\usepackage{color}
\usepackage{amsfonts}
\usepackage{multirow}
\usepackage{diagbox}
\usepackage{bm}
\usepackage{fdsymbol}

\newcommand{\ie}{\textit{i}.\textit{e}.}
\newcommand{\eg}{\textit{e}.\textit{g}.}

\newlength\savewidth\newcommand\shline{\noalign{\global\savewidth\arrayrulewidth\global\arrayrulewidth 1pt}\hline\noalign{\global\arrayrulewidth\savewidth}}

\title{Unleashing the Power of Prompt-driven Nucleus Instance Segmentation}

\author{
	Anonymous
}

\author{
	Zhongyi Shui$^{1,3}$
	\and
	Yunlong Zhang$^{1,3}$\and
	Kai Yao$^{2}$\and
	Chenglu Zhu$^{3}$\and
	Sunyi Zheng$^{3}$\and \\
	Jingxiong Li$^{1,3}$\and
	Honglin Li$^{1,3}$\and
	Yuxuan Sun$^{1,3}$\and
	Ruizhe Guo$^{1,3}$\and
	Lin Yang$^{3}$
	\affiliations
	$^1$College of Computer Science and Technology, Zhejiang University\\
	$^2$University of Liverpool\\
	$^3$School of Engineering, Westlake University\\
	\emails
	\{shuizhongyi, yanglin\}@westlake.edu.cn
}

\begin{document}
	
\maketitle

\begin{abstract}
	Nucleus instance segmentation in histology images is crucial for a broad spectrum of clinical applications. Current dominant algorithms rely on regression of nuclear proxy maps. Distinguishing nucleus instances from the estimated maps requires carefully curated post-processing, which is error-prone and parameter-sensitive. Recently, the Segment Anything Model (SAM) has earned huge attention in medical image segmentation, owing to its impressive generalization ability and promptable property.  Nevertheless, its potential on nucleus instance segmentation remains largely underexplored. In this paper, we present a novel prompt-driven framework that consists of a nucleus prompter and SAM for automatic nucleus instance segmentation. Specifically, the prompter learns to generate a unique point prompt for each nucleus while the SAM is fine-tuned to output the corresponding mask for the prompted nucleus. Furthermore, we propose the inclusion of adjacent nuclei as negative prompts to enhance the model's capability to identify overlapping nuclei. Without complicated post-processing, our proposed method sets a new state-of-the-art performance on three challenging benchmarks. Code is available at \url{github.com/windygoo/PromptNucSeg}
\end{abstract}

\section{Introduction}
Cancer is one of the leading causes of death worldwide. Over the past decades, substantial endeavors have been made to detect cancers from histology images with the aim of improving survival rates through early screening. Identification of nuclear components in the histology landscape is often the first step toward a detailed analysis of histology images. Quantitative characterizations of nuclear morphology and structure play a pivotal role in cancer diagnosis, treatment planning, and survival analysis, which have been verified by a wide range of studies, see for example \cite{alberts2015essential}. However, large-scale analysis on the cell level is extremely labor-intensive and time-consuming since a whole slide image (WSI) typically contains tens of thousands of nuclei of various types. Moreover, such subjective interpretations have been demonstrated to suffer from large inter-and intra-observer variability \cite{he2021cdnet}. Consequently, there is a compelling pursuit of precise automatic algorithms for nucleus instance segmentation to aid in histopathologic cancer diagnosis. Nonetheless, the blurred cell contours, overlapping cell clusters, and variances in nuclei staining, shape and size, pose substantial challenges for the developers.
\begin{figure}[t!]
	\begin{center}		
		\includegraphics[width=0.98\linewidth]{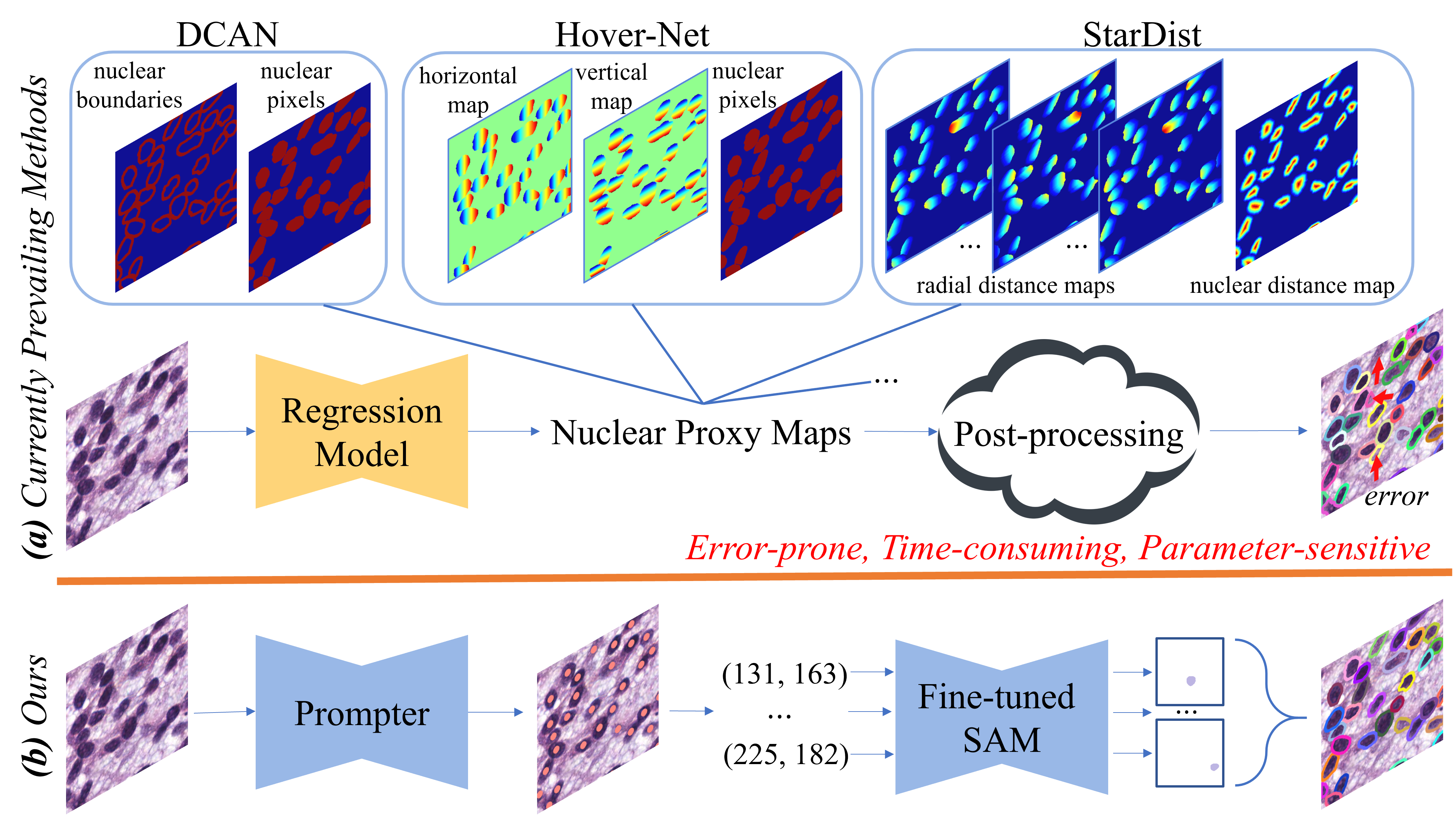}
		\caption{Pipeline comparison with currently prevailing nucleus instance segmentation algorithms.}
		\label{fig:banner}
		\vspace{-15pt}
	\end{center}
\end{figure}

Recent years have witnessed significant advancements in the filed of nucleus instance segmentation owing to the impressive performances brought by various methods based on regression of nuclear proxy maps \cite{chen2016dcan,naylor2018segmentation,schmidt2018cell,zhou2019cia,graham2019hover,zhou2019cia,ilyas2022tsfd,chen2023cpp,chen2023enhancing} (see Fig.~\ref{fig:banner} (a)). Regrettably, these methods necessitate carefully crafted post-processing to derive nuclear instances from the estimated maps. This step demands meticulous hyper-parameter tuning and is vulnerable to noise \cite{yao2023pointnu}.

Recently, the segment anything model (SAM) has emerged as a generic segmentation network for various image types, whose impressive generalization ability and versatility can be attributed to its structural design and the strong representation learned from 11M images annotated with 1B masks \cite{kirillov2023segment}. Several studies have been undertaken to investigate the zero-shot performance of SAM on nucleus segmentation \cite{deng2023segment} or transfer its well-learned representation to boost the segmentation accuracy \cite{horst2023cellvit,xu2023sppnet}. Specifically, \cite{horst2023cellvit} reuses SAM's well-trained image encoder to construct a more powerful regression model and integrates it into the aforementioned nucleus instance segmentation workflow. Despite the promising results, we argue that this approach does not fully exploit the knowledge encapsulated in the integrated architecture of SAM. Conversely, \cite{xu2023sppnet} maintains the philosophy of SAM thoroughly. They fine-tune the entire SAM in a one-prompt-all-nuclei recipe for nucleus semantic segmentation. Nevertheless, this method expects users to supply precise prompts, which is impractical since crafting such prompts requires extensive medical expertise. Moreover, it falls short in providing nucleus instance information.

In this work, we propose to fine-tune SAM in a one-prompt-one-nucleus regime to fully unleash its potential for nucleus instance segmentation. To eliminate the need for crafted prompts during inference, we develop a prompter that automatically generates nuclei prompts by refining and classifying pre-defined anchor points on an input image. Specially, we incorporate an auxiliary task of nuclear region segmentation into prompter learning. This integration guides the model's attention towards foreground areas, thereby improving the quality of generated prompts. During the inference stage, the predicted nuclear region mask is further utilized to filter out false positive prompts. The consolidation of the prompter and segmentor (i.e., the fine-tuned SAM) establishes a novel solution for automatic nucleus instance segmentation. Given their linkage through nuclei prompts, we designate our approach as PromptNucSeg, and its pipeline is depicted in Fig.~\ref{fig:banner} (b). Compared to the currently prevailing methods, our approach does not require complex post-processing. Moreover, we devise a trick that treats adjacent nuclei as negative prompts to improve the model's segmentation of overlapping nuclei.

Our contributions can be summarized as follows:
\begin{itemize}
	\item We propose PromptNucSeg, which provides a new perspective for nucleus instance segmentation.
	\item We develop a prompter for automatic nuclei prompts generation and design a simple auxiliary task to boost its performance.
	\item We propose to use adjacent nuclei as negative prompts to promote segmentation of overlapping nuclei.
	\item Extensive experiments on three challenging benchmarks demonstrate the advantages of PromptNucSeg over the state-of-the-art counterparts.	
\end{itemize}

\section{Related Work}
\subsection{Utilization of SAM for Medical Image segmentation}
Segment Anything Model (SAM) \cite{kirillov2023segment} is the first groundbreaking model for universal image segmentation. It has achieved impressive results on a wide range of natural image tasks. Nevertheless, due to the dramatic domain gap between natural and medical images, SAM’s performance significantly declines when applied for medical image segmentation \cite{huang2023segment,ma2023segment}. To bridge this gap, many studies opt to fine-tine SAM with meticulously curated medical data \cite{cheng2023sam,wang2023sam,wu2023medical,zhang2023customized,lin2023samus,lei2023medlsam}. These works mainly focus on the segmentation of anatomical structures and organs in computed tomography (CT), magnetic resonance (MR) and ultrasound images.

In terms of histology images, \cite{deng2023segment} assesses SAM's performance for tumor, non-tumor tissue and nucleus segmentation. The results suggest that the vanilla SAM achieves remarkable segmentation performance for large connected tissue objects, however, it does not consistently achieve satisfactory results for dense nucleus instance segmentation. To tackle this issue, SPPNet \cite{xu2023sppnet} fine-tunes a distilled lightweight SAM \cite{zhang2023faster} in a one-prompt-all-nuclei manner for nucleus semantic segmentation. Despite the improved outcomes, this method relies on manual prompts and lacks the capacity to furnish nucleus instance information. \cite{horst2023cellvit} builds a vision transformer-based U-Net-shaped model, employing SAM's pre-trained image encoder as its backbone to better fit the nuclear proxy maps. We argue that this approach underutilizes the knowledge embedded in SAM's integrated architecture.

\subsection{Nucleus Instance Segmentation}
Current methods for nucleus instance segmentation can be divided into two categories: top-down and bottom-up.

Top-down methods, such as Mask R-CNN \cite{he2017mask}, first predict nuclei bounding boxes from a global perspective, and then segment the nucleus instance within each box. Despite the great progress in natural image segmentation and the potential in dealing with overlapping nuclei, top-down methods have demonstrated deficiency on nucleus instance segmentation \cite{graham2019hover,yao2023pointnu,lou2023structure}, attributed to two primary factors. First, on the data side, there are many severely overlapping nuclei in histology images. Consequently, a bounding-box proposal normally contains multiple nuclei with indistinct boundaries, making the network hard to optimize. Second, on the model side, top-down methods typically predict segmentation masks with a fixed resolution (e.g., 28$\times$28 in Mask R-CNN). Subsequently, these masks undergo re-sampling to match the size of their corresponding bounding boxes. This re-sampling process might introduce quantization errors \cite{yao2023pointnu}, posing challenges for accurately segmenting sinuous nuclear boundaries.
\begin{figure*}[t!]
	\begin{center}
		\includegraphics[width=1.0\linewidth]{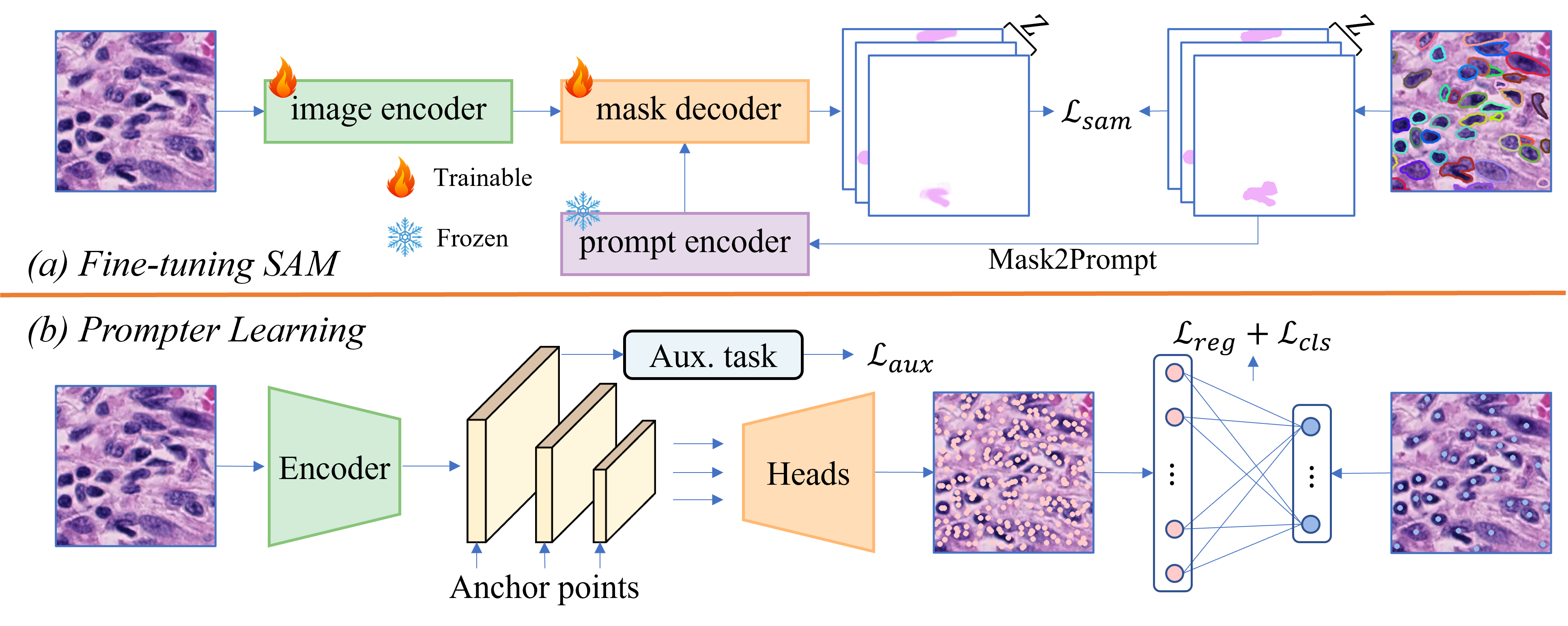}
		\vspace{-15pt}
		\caption{(a) The fine-tuning process of SAM. Mask2Prompt signifies randomly sampling a positive point prompt from the foreground area of each nucleus mask. (b) The training procedure of the nucleus prompter. The integration of these two models enables automatic nucleus instance segmentation, as illustrated in Fig.~\ref{fig:banner} (b).}
		\label{fig:framework}
		\vspace{-15pt}
	\end{center}
\end{figure*}

Bottom-up methods, initially regressing various types of nuclear proxy maps and then grouping pixels into individual instances through meticulous post-processing, have gained prominence in nucleus instance segmentation owing to their commendable accuracy. These approaches typically entail regressing a nucleus probability map, where the pixel values signify the presence of nuclei, along with some auxiliary maps facilitating the identification of nuclei instances. Specifically, DCAN \cite{chen2016dcan}, CIA-Net \cite{zhou2019cia}, TSFD-Net \cite{ilyas2022tsfd} and HARU-Net \cite{chen2023enhancing} predict the nuclear contour map. DIST \cite{naylor2018segmentation} regresses the intra-nuclear distance map. HoVer-Net \cite{graham2019hover} predicts horizontal and vertical distances of nuclei pixels to their center of mass. StarDist \cite{schmidt2018cell} and its extension CPP-Net \cite{chen2023cpp} predict distances from each foreground pixel to its associated instance boundary along a set of pre-defined directions. Under the premise of some above frameworks, other works \cite{qu2019improving,zhao2020triple,deshmukh2022feednet,horst2023cellvit} put effort into constructing more favorable features or task-specific loss functions. Overall, while bottom-up methods have exhibited superior accuracy compared to top-down approaches, their accompanying post-processing requires tedious hyper-parameter tuning \cite{yao2023pointnu}, which presents a hurdle to their practical application.



Essentially, our proposed PromptNucSeg belongs the top-down family. But inspired by the promptable property of SAM, we tackle this task from a new perspective. Instead of bounding boxes, we utilize center points to represent nuclei, which are easier to localize and can separate touching objects more precisely. In comparison with bottom-up methods, PromptNucSeg does not require intricate post-processing as the prompter generates point prompts for nuclei in a one-to-one relationship and the segmentor predicts the nuclei mask guided by each prompt individually.

\section{Methodology}
\subsection{Preliminaries: SAM}
SAM \cite{kirillov2023segment} consists of three sub-networks, \ie, image encoder $\mathcal{F}$, prompt encoder $\mathcal{P}$ and mask decoder $\mathcal{M}$. The image encoder transforms an input image $I\in \mathbb{R}^{H\times W\times3}$ into an image embedding. The prompt encoder maps diverse prompts (\eg, a set of positive/negative points, a rough box or mask, free-form text, or combinations thereof) into a compact prompt embedding. Positive prompts indicate regions representing the region-of-interest (ROI) object, whereas negative prompts emphasize areas that should be suppressed as background. Given the image and prompt embedding as input, the mask decoder generates the mask for the ROI object in conjunction with a confidence score (\ie, an estimated IoU).

\subsection{Adapt SAM to nucleus instance segmentation}
Despite SAM's remarkable segmentation performance across numerous natural images, recent studies have highlighted its subpar performance on medical images due to the significant domain gap \cite{cheng2023sam,huang2023segment}. A specific observation worth noting is that the objects in SAM's pre-training data are primarily captured in natural scenes, displaying nicely delineated boundaries, while the boundaries of organs or nuclei in medical images are often ambiguous \cite{cheng2023sam,xu2023sppnet}. To enhance the capability of SAM for nucleus segmentation, we fine-tune it on nucleus instance segmentation datasets to incorporate essential domain-specific knowledge into the model. 

The fine-tuning procedure is depicted in Fig.~\ref{fig:framework} (a). Specifically, for each image-label pair $(x,y)$ in a mini-batch, we randomly select $Z$ nucleus instances from the instance map $y$. Subsequently, a positive point prompt is randomly sampled from the foreground area of each instance. Taking the image $x$ and the point prompt $p_z$ as input, we fine-tune SAM to predict the mask of $z$-th nucleus instance.
\begin{equation}\label{eq:forward}
	\widetilde{\mathcal{O}}_z=\mathcal{M}\left(\mathcal{F}\left(x\right),\mathcal{P}\left(\{p_z\}\right),[\text{mask}],[\text{IoU}]\right)
\end{equation}
where $[\text{mask}]$ and $[\text{IoU}]$ separately represent the learnable mask and IoU token pre-set in SAM's mask decoder. $\widetilde{\mathcal{O}}_z$ denotes the predicted mask of the $z$-th nucleus. We supervise the mask and IoU prediction with the same loss as SAM.
\begin{equation}\label{eq:loss}
	\mathcal{L}_{sam}=\omega\text{FL}\left(\widetilde{\mathcal{O}}_z,\mathcal{O}_z\right) + \text{DL}\left(\widetilde{\mathcal{O}}_z,\mathcal{O}_z\right)+\text{MSE}\left(\widetilde{\nu},\nu\right)
\end{equation}
where FL, DL and MSE stand for focal loss \cite{lin2017focal}, dice loss \cite{milletari2016v} and mean-square-error loss, respectively. $\mathcal{O}_z$ is the ground-truth mask of the $z$-th nucleus, $\widetilde{\nu}$ and $\mathcal{\nu}$ signify the estimated and actual IoU between $\widetilde{\mathcal{O}_z}$ and $\mathcal{O}_z$, respectively. $\omega$ is a weight term. In this work, we opt to freeze the prompt encoder while updating the image encoder and mask decoder via gradient descent.

\subsection{Learn Prompter}
Generating a unique point prompt for each nucleus is de facto a non-trivial problem. In this study, we choose the nuclear centroid as its prompt for simplicity. To achieve automatic prompt generation, we draw inspiration from \cite{song2021rethinking} and develop a prompter to predict nuclear centroid coordinates and categories by refining and classifying a set of anchor points placed on an input image. In the following content, we denote the set of anchor points as $\mathcal{A}=\{a_i\}_{i=1}^{M}$ and the set of ground-truth points as $\mathcal{B}=\{b_i\}_{i=1}^{N}$, where $b_i$ is extracted from $y$ as the centroid of the $i$-th nucleus.

The prompter learning procedure is depicted in Fig.~\ref{fig:framework} (b). Specifically, we begin with placing anchor points on an input image $x$ with a step of $\lambda$ pixels. Then, an image encoder $\mathcal{F}^\prime$ is employed to construct hierarchical feature maps $\{P_j\}_{j=2}^L$ from $x$, where the size of $P_j$ is $(H/2^j,W/2^j)$. Following this, we apply the bilinear interpolation method to extract the multi-scale feature vectors $\{f_{i,j}\}_{j=2}^L$ for anchor point $a_i$ according to its normalized coordinates on the feature pyramid. Finally, we concatenate $\{f_{i,j}\}_{j=2}^L$ and fed it into two dedicated MLP heads for decoding offsets $\delta_i$ and logits $q_i\in\mathbb{R}^{C+1}$ with respect to $a_i$, where $C$ is the number of nuclear categories and the extra class is background.

Since the goal of prompter is to associate a unique point prompt for each nucleus, which anchor point in $\mathcal{A}$ should be chosen as the prompt is the key in prompter learning. In principal, for any nucleus centroid in $\mathcal{B}$, the anchor point with lower distance and higher categorical similarity with it is preferred to be chosen. Consequently, the association can be completed by computing the maximum-weight matching $\phi=\{(a_{\sigma(i)},b_i)\}_{i=1}^N$ in a weighted bipartite graph $\mathcal{G}=(\mathcal{A},\mathcal{B},\mathcal{E})$, where  the weight $w_{i,j}$ of edge connecting vertex $a_i$ and $b_j$ is defined as:
\begin{equation} 
	w_{i,j}=q_i(c_j)-\alpha||\hat{a}_i-b_j||_2
\end{equation}
in which $c_j$ is the class of the $j$-th nucleus, $\hat{a}_i=a_i+\delta_i$ represents the refined position of the $i$-th anchor point, $q_i(c_j)$ is the $c_j$-th element of $q_i$, $\alpha$ is a weight term and $||\cdot||_2$ denotes $l_2$ distance. We use the Hungarian algorithm \cite{song2021rethinking} to determine $\phi$ in this work. As a result, the objective of prompter is concretized as narrowing the positional and categorical difference between the selected anchor points and their matched nuclei, while ignoring the unselected anchor points as background. This objective can be achieved by minimizing the following losses.
\begin{equation}
	\begin{aligned}
		\mathcal{L}_{\text{cls}} &= -\frac{1}{M}\left(\sum_{i=1}^N\log q_{\sigma(i)}\left(c_i\right) + \beta\sum_{a_i\in\mathcal{A}^\prime}\log q_{i}\left(\varnothing\right)\right) \\ 
		\mathcal{L}_{\text{reg}} &=  \frac{\gamma}{N}\sum_{i=1}^N||\hat{a}_{\sigma(i)} - b_i||_2
	\end{aligned}
\end{equation}
where $\mathcal{A}^\prime\subsetneqq\mathcal{A}$ represents the set of unselected anchor points, $\varnothing$ indicates the background class, $\beta$ and $\gamma$ are free parameters used to relieve the class imbalance and modulate the effect of regression loss, respectively.

\paragraph{Auxiliary task of nuclear region segmentation}
The training process of the above prompter only involves the nuclear categorical labels and centroid coordinates. However, in the context of nucleus instance segmentation, the mask for each nucleus is also available, which provides rich details about nuclear size, shape and so on. To integrate this valuable information into prompter learning, we construct a simple auxiliary task of nuclear region segmentation to enhance the model's attention to foreground areas and   perception of nuclear morphological characteristics. Technically, we introduce a mask head structured as Conv-BN-ReLu-Conv to predict the nuclear probability map $\hat{S}$ from $P_2$, informed by that the high-resolution $P_2$ contains abundant fine-grained features crucial for medical image segmentation \cite{lin2017feature}. We apply the focal loss to supervise the learning of the auxiliary task.
\begin{equation}
	\begin{aligned}
		\mathcal{L}_{\text{aux}} = \text{FL}\left(\hat{S},S\right)
	\end{aligned}
\end{equation}
where the ground-truth mask $S$ is derived from the instance map $y$ via a simple thresholding operation. The final loss used to optimize the prompter is
\begin{equation}
	\begin{aligned}
		\mathcal{L}_{\text{prompter}} = \mathcal{L}_{\text{reg}} + \mathcal{L}_{\text{cls}} + \mathcal{L}_{\text{aux}}
	\end{aligned}
\end{equation}

\paragraph{Mask-aided prompt filtering}
Due to insufficient optimization, the prompter would inevitably produce false positive prompts that actually represent non-nucleus objects. To mitigate this issue, we utilize the nuclear probability map predicted by the auxiliary branch to filter out these incorrect predictions. This is achieved by retaining only those prompts with probability values exceeding 0.5 in the inference stage.

\subsection{Use adjacent nuclei as negative prompts}
\begin{figure}[t!]
	\centering		
	\includegraphics[width=0.98\linewidth]{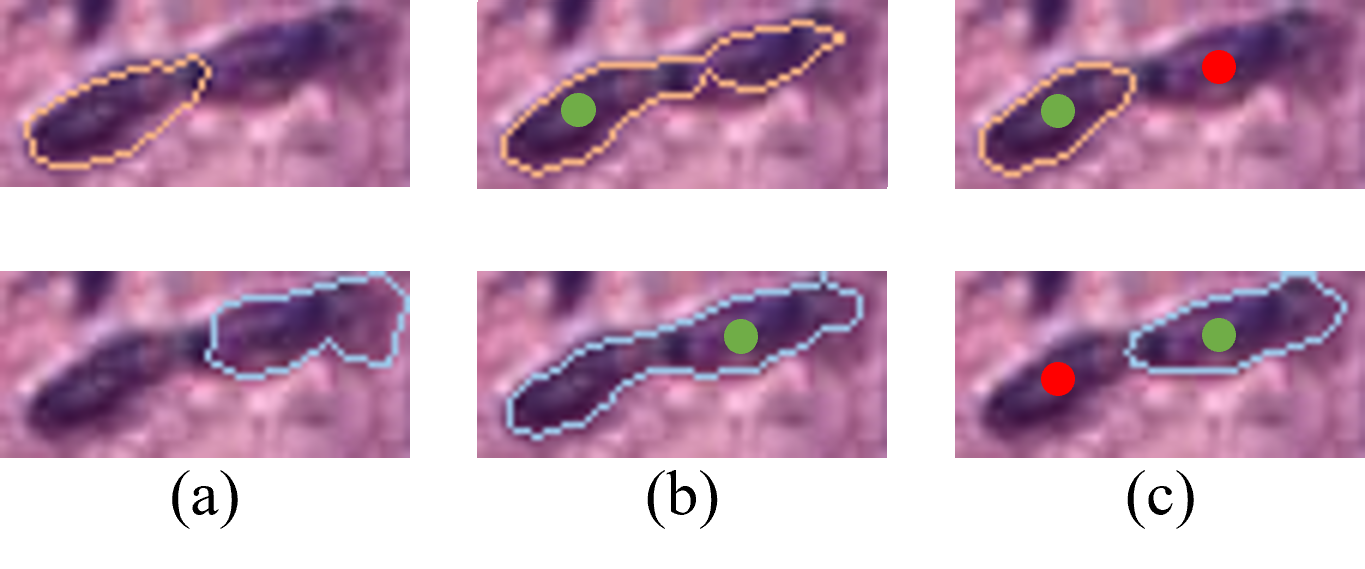}
	\vspace{-15pt}
	\caption{(a) Ground-truth boundary of two overlapping nuclei. (b) Predicted boundary by prompting each nucleus with a positive prompt inside it. (c) Predicted boundary by prompting each nucleus with an additional negative prompt inside its overlapping nucleus. \textcolor{green}{$\mdlgblkcircle$} Positive prompt	\textcolor{red}{$\mdlgblkcircle$} Negative prompt}
	\label{fig:np}
\end{figure}

Distinguishing overlapping nuclei is a long-standing challenge in the community of nucleus instance segmentation \cite{graham2019hover,horst2023cellvit,he2023toposeg}. Our approach encounters this challenge as well. Given a fine-tuned SAM, considering a real-world scenario of two overlapping nuclei in Fig.~\ref{fig:np} (a), prompting each nucleus with a single positive prompt results in an over-segmented mask due to the faint boundary, as depicted in Fig.~\ref{fig:np} (b). An intuitive idea to resolve this problem is to include the overlapping nucleus as negative prompt to suppress excessive segmentation for the ROI nucleus, as illustrated in Fig.~\ref{fig:np} (c).

Nevertheless, the implementation of this idea presents two practical challenges. (1) In the inference phase, it is unknown which nuclei overlap with a ROI nucleus. (2) We empirically observe that including negative prompts solely at test time cannot effectively prevent over-segmented prediction for overlapping nuclei. The inefficiency stems from that the fine-tuning process involving only positive prompts (see Eq.~\ref{eq:forward}) causes a catastrophic forgetting about the effect of negative prompts.

To deal with (1), let $\hat{p}_z$ denote the generated point prompt for the $z$-th nucleus in a test image, we approximately employ the $K$ points nearest to $\hat{p}_z$ as negative prompts for segmenting this nucleus. To address (2), we incorporate negative prompts into the fine-tuning stage in a similar way. Specifically, we randomly sample a point from each nucleus instance in $y$ and utilize the positive prompt $p_z$ along with its $K$-nearest points $\{n_{z,k}\}_{k=1}^K$ as negative prompts to predict the mask of the $z$-th nucleus. As a result, we re-formulate the model's forward process described by Eq.~\ref{eq:forward} as
\begin{equation}\label{eq:forward1}
	\widetilde{\mathcal{O}}_z=\mathcal{M}\left(\mathcal{F}\left(x\right),\mathcal{P}\left(\{p_z\}\cup\{n_{z,k}\}_{k=1}^K\right),[\text{mask}],[\text{IoU}]\right)
\end{equation}

\subsection{Inference}
In line with \cite{lin2023samus,cheng2023sam}, we scale down the input spatial resolution of SAM from 1024$\times$1024 to 256$\times$256 for more clinical-friendly deployment of our method. This adaption brings a substantial reduction in GPU memory due to the shorter input sequence in attention layers.

Given an image of arbitrary resolution in the inference stage, we first employ the prompter to predict nuclei prompts from a global view. If the longer side of the image is more than 256 pixel, we do not directly apply the segmentor on it to avoid the performance degradation resulting from the interpolation of positional embedding \cite{su2021roformer}. Instead, we partition the image into tiles of size 256$\times$256 and gather the masks predicted within each local view. To ensure any nucleus can appear completely in at least one tile, the partitioning is performed in a sliding window manner modulated by an overlapping size of $\epsilon$. In the end, we apply a simple non-maximum suppression to eliminate duplicate predictions.

\begin{table*}[t!]
	\centering
	\small{
		\resizebox{0.99\linewidth}{!}{		
			\begin{tabular}{c|cc|cc|cc|cc|cc|cc|cc}
				\toprule[1.5pt]
				\multirow{3}{*}{Tissue} & \multicolumn{2}{c|}{Mask R-CNN} &
				\multicolumn{2}{c|}{StarDist} & \multicolumn{2}{c|}{Hover-Net} & \multicolumn{2}{c|}{CPP-Net} & \multicolumn{2}{c|}{PointNu-Net} & \multicolumn{2}{c|}{CellViT-H} &
				\multicolumn{2}{c}{PromptNucSeg-H} \\
				& \multicolumn{2}{c|}{\cite{he2017mask}} &
				\multicolumn{2}{c|}{\cite{schmidt2018cell}} & \multicolumn{2}{c|}{\cite{graham2019hover}} & \multicolumn{2}{c|}{\cite{chen2023cpp}}& \multicolumn{2}{c|}{\cite{yao2023pointnu}} & \multicolumn{2}{c|}{\cite{horst2023cellvit}} & \multicolumn{2}{c}{(Ours)} \\
				\cmidrule(lr){2-3} \cmidrule(lr){4-5} \cmidrule(lr){6-7} \cmidrule(lr){8-9} \cmidrule(lr){10-11} \cmidrule(lr){12-13} \cmidrule(lr){14-15}
				& bPQ & mPQ & bPQ & mPQ & bPQ & mPQ & bPQ & mPQ & bPQ & mPQ & bPQ & mPQ & bPQ & mPQ \\
				\shline
				Adrenal & 0.5546 & 0.3470 & 0.6972 & 0.4868 & 0.6962 & 0.4812 & 0.7066 & 0.4944 & \underline{0.7134} & 0.5115 & 0.7086 & \textbf{0.5134}             & \textbf{0.7227} & \underline{0.5128} \\
				Bile Duct      & 0.5567 & 0.3536 & 0.6690 & 0.4651 & 0.6696 & 0.4714 & 0.6768 & 0.4670 & \underline{0.6814} & 0.4868 & 0.6784 & \underline{0.4887}             & \textbf{0.6976} & \textbf{0.5012} \\
				Bladder        & 0.6049 & 0.5065 & 0.6986 & 0.5793 & 0.7031 & 0.5792 & 0.7053 & 0.5936 & \textbf{0.7226} & \textbf{0.6065} & 0.7068 & 0.5844             & \underline{0.7212} & \underline{0.6043} \\
				Breast         & 0.5574 & 0.3882 & 0.6666 & 0.5064 & 0.6470 & 0.4902 & 0.6747 & 0.5090 & 0.6709 & 0.5147 & \underline{0.6748} & \underline{0.5180}             & \textbf{0.6842} & \textbf{0.5322} \\
				Cervix         & 0.5483 & 0.3402 & 0.6690 & 0.4628 & 0.6652 & 0.4438 & \underline{0.6912} & 0.4792 & 0.6899 & \underline{0.5014} & 0.6872 & 0.4984             & \textbf{0.6983} & \textbf{0.5118} \\
				Colon          & 0.4603 & 0.3122 & 0.5779 & 0.4205 & 0.5575 & 0.4095 & 0.5911 & 0.4315 & \underline{0.5945} & \underline{0.4509} & 0.5921 & 0.4485             & \textbf{0.6096} & \textbf{0.4690} \\
				Esophagus      & 0.5691 & 0.4311 & 0.6655 & 0.5331 & 0.6427 & 0.5085 & \underline{0.6797} & 0.5449 & 0.6766 & \underline{0.5504} & 0.6682 & 0.5454 & \textbf{0.6920}  & \textbf{0.5711} \\
				Head \& Neck   & 0.5457 & 0.3946 & 0.6433 & 0.4768 & 0.6331 & 0.4530 & 0.6523 & 0.4706 & \underline{0.6546} & 0.4838 & 0.6544 & \underline{0.4913} & \textbf{0.6695} & \textbf{0.5104} \\
				Kidney         & 0.5092 & 0.3553 & 0.6998 & 0.4880 & 0.6836 & 0.4424 & 0.7067 & 0.5194 & 0.6912 & 0.5066 & \underline{0.7092} & \underline{0.5366}             & \textbf{0.7115} & \textbf{0.5786} \\
				Liver          & 0.6085 & 0.4103 & 0.7231 & 0.5145 & 0.7248 & 0.4974 & 0.7312 & 0.5143 & 0.7314 & 0.5174 & \underline{0.7322} & \underline{0.5224}             & \textbf{0.7372} & \textbf{0.5333} \\
				Lung           & 0.5134 & 0.3182 & 0.6362 & 0.4128 & 0.6302 & 0.4004 & 0.6386 & 0.4256 & 0.6352 & 0.4048 & \underline{0.6426} & \underline{0.4314} & \textbf{0.6580}  & \textbf{0.4398} \\
				Ovarian        & 0.5784 & 0.4337 & 0.6668 & 0.5205 & 0.6309 & 0.4863 & 0.6830 & 0.5313 & \textbf{0.6863} & \textbf{0.5484} & 0.6722 & 0.5390             & \underline{0.6856} & \underline{0.5442} \\
				Pancreatic     & 0.5460 & 0.3624 & 0.6601 & 0.4585 & 0.6491 & 0.4600 & 0.6789 & 0.4706 & \underline{0.6791} & \underline{0.4804} & 0.6658 & 0.4719             & \textbf{0.6863} & \textbf{0.4974} \\
				Prostate       & 0.5789 & 0.3959 & 0.6748 & 0.5067 & 0.6615 & 0.5101 & \underline{0.6927} & 0.5305 & 0.6854 & 0.5127 & 0.6821 & \underline{0.5321}             & \textbf{0.6983} & \textbf{0.5456} \\
				Skin           & 0.5021 & 0.2665 & 0.6289 & 0.3610 & 0.6234 & 0.3429 & 0.6209 & 0.3574 & 0.6494 & 0.4011 & \underline{0.6565} & \textbf{0.4339}             & \textbf{0.6613} & \underline{0.4113} \\
				Stomach        & 0.5976 & 0.3684 & 0.6944 & 0.4477 & 0.6886 & \textbf{0.4726} & \underline{0.7067} & 0.4582 & 0.7010 & 0.4517 & 0.7022 & \underline{0.4705}             & \textbf{0.7115} & 0.4559 \\
				Testis         & 0.5420 & 0.3512 & 0.6869 & 0.4942 & 0.6890 & 0.4754 & 0.7026 & 0.4931 & \underline{0.7058} & \underline{0.5334} & 0.6955 & 0.5127             & \textbf{0.7151} & \textbf{0.5474} \\
				Thyroid        & 0.5712 & 0.3037 & 0.6962 & 0.4300 & 0.6983 & 0.4315 & \underline{0.7155} & 0.4392 & 0.7076 & 0.4508 & 0.7151 & \underline{0.4519}             & \textbf{0.7218} & \textbf{0.4721} \\
				Uterus         & 0.5589 & 0.3683 & 0.6599 & 0.4480 & 0.6393 & 0.4393 & 0.6615 & 0.4794 & \underline{0.6634} & \underline{0.4846} & 0.6625 & 0.4737             & \textbf{0.6743} & \textbf{0.4955} \\
				\shline
				Average & 0.5528 & 0.3688 & 0.6692 & 0.4744 & 0.6596 & 0.4629 & 0.6798 & 0.4847 & \underline{0.6808} & 0.4957 & 0.6793 & \underline{0.4980} & \textbf{0.6924} & \textbf{0.5123} \\
				Std & 0.0076 & 0.0047 & 0.0014 & 0.0037 & 0.0036 & 0.0076 & 0.0015 & 0.0059 & 0.0050 & 0.0082 & 0.0318 & 0.0413 & 0.0093 & 0.0147 \\
				\bottomrule[1.5pt]
	\end{tabular}}}
	\caption{Performance comparison on the PanNuke dataset. Following \protect\cite{chen2023cpp,horst2023cellvit}, both binary PQ (bPQ) and multi-class PQ (mPQ) are computed for evaluation. The best and second-best PQ scores are highlighted in \textbf{bold} and \underline{underlined}.}
	\label{tab:pannuke}
	\vspace{-5pt}
\end{table*}

\begin{table*}[t!]
	\centering
	\resizebox{0.98\linewidth}{!}{		
		\begin{tabular}{ccccccccccccccccccc}
			\shline
			\multirow{3}{*}{Method} & \multicolumn{3}{c}{\multirow{2}{*}{Detection}} & \multicolumn{15}{c}{Classification} \\
			\cmidrule{5-19}
			& & & &
			\multicolumn{3}{c}{Neoplastic} &
			\multicolumn{3}{c}{Epithelial} &
			\multicolumn{3}{c}{Inflammatory} &
			\multicolumn{3}{c}{Connective} &
			\multicolumn{3}{c}{Dead} \\
			\cmidrule(lr){2-4} \cmidrule(lr){5-7} \cmidrule(lr){8-10} \cmidrule(lr){11-13} \cmidrule(lr){14-16} \cmidrule(lr){17-19}
			& P & R & F1 & P & R & F1 & P & R & F1 & P & R & F1 & P & R & F1 & P & R & F1 \\
			\shline
			Mask-RCNN \cite{he2017mask} & 0.76 & 0.68 & 0.72 & 0.55 & 0.63 & 0.59 & 0.52 & 0.52 & 0.52 & 0.46 & 0.54 & 0.50 & 0.42 & 0.43 & 0.42 & 0.17 & 0.30 & 0.22 \\
			DIST \cite{naylor2018segmentation} & 0.74 & 0.71 & 0.73 & 0.49 & 0.55 & 0.50 & 0.38 & 0.33 & 0.35 & 0.42 & 0.45 & 0.42 & 0.42 & 0.37 & 0.39 & 0.00 & 0.00 & 0.00 \\
			StarDist \cite{schmidt2018cell} & 0.85 & 0.80 & 0.82 & 0.69 & 0.69 & 0.69 & 0.73 & 0.68 & 0.70 & 0.62 & 0.53 & 0.57 & 0.54 & 0.49 & 0.51 & 0.39 & 0.09 & 0.10 \\
			Micro-Net \cite{raza2019micro} & 0.78 & 0.82 & 0.80 & 0.59 & 0.66 & 0.62 & 0.63 & 0.54 & 0.58 & 0.59 & 0.46 & 0.52 & 0.50 & 0.45 & 0.47 & 0.23 & 0.17 & 0.19 \\
			Hover-Net \cite{graham2019hover} & 0.82 & 0.79 & 0.80 & 0.58 & 0.67 & 0.62 & 0.54 & 0.60 & 0.56 & 0.56 & 0.51 & 0.54 & 0.52 & 0.47 & 0.49 & 0.28 & 0.35 & 0.31 \\
			CPP-Net \cite{chen2023cpp} & 0.87 & 0.78 & 0.82 & 0.74 & 0.67 & \underline{0.70} & 0.74 & 0.70 & 0.72 & 0.60 & 0.57 & \underline{0.58} & 0.57 & 0.49 & \underline{0.53} & 0.41 & 0.36 & \underline{0.38} \\
			CellViT-H \cite{horst2023cellvit} & 0.84 & 0.81 & \underline{0.83} & 0.72 & 0.69 & \textbf{0.71} & 0.72 & 0.73 & \underline{0.73} & 0.59 & 0.57 & \underline{0.58} & 0.55 & 0.52 & \underline{0.53} & 0.43 & 0.32 & 0.36 \\
			PromptNucSeg & 0.82 & 0.85 & \textbf{0.84} & 0.70 & 0.72 & \textbf{0.71} & 0.73 & 0.78 & \textbf{0.76} & 0.58 & 0.61 & \textbf{0.59} & 0.55 & 0.55 & \textbf{0.55} & 0.44 & 0.49 & \textbf{0.46} \\
			\shline
	\end{tabular}}
	\caption{Precision (P), Recall (R) and F1-score (F1) for detection and classification across three folds for each nucleus type. The best F1-score is in \textbf{bold} while the second best is \underline{underlined}. Following \protect\cite{graham2019hover}, if a detected nucleus is within a valid distance ($\approx$3$\mu$m) from an annotated nucleus and the nuclear class matches, it is counted as a true positive (TP), otherwise a false positive(FP).}
	\label{tab:pannuke_det}
	\vspace{-9pt}
\end{table*}

\begin{table}[t!]
	\centering
	\resizebox{0.99\linewidth}{!}{		
		\begin{tabular}{c|ccccc}
			\shline
			\diagbox[width=8em]{{Method}}{{Class}} & Neoplastic & Epithelial & Inflammatory & Connective & Dead \\
			\shline
			Mask-RCNN & 0.472 & 0.403 & 0.290 & 0.300 & 0.069 \\
			DIST & 0.439 & 0.290 & 0.343 & 0.275 & 0.000 \\
			StarDist & 0.547 & 0.532 & 0.424 & 0.380 & 0.123 \\
			Micro-Net & 0.504 & 0.442 & 0.333 & 0.334 & 0.051 \\
			HoVer-Net & 0.551 & 0.491 & 0.417 & 0.388 & 0.139 \\
			CPP-Net & 0.571 & 0.565 & 0.405 & 0.395 & 0.131 \\
			PointNu-Net & 0.578 & 0.577 & \underline{0.433} & 0.409 & \underline{0.154} \\
			CellViT-H & \underline{0.581} & \textbf{0.583} & 0.417 & \underline{0.423} & 0.149 \\
			PromptNucSeg-H & \textbf{0.598} & \underline{0.582} & \textbf{0.441} & \textbf{0.433} & \textbf{0.161} \\
			\shline
	\end{tabular}}
	\caption{Average PQ across three folds for each nuclear category on the PanNuke dataset.}
	\vspace{-10pt}
	\label{tab:pannuke_seg}
\end{table}

\section{Experiment}

\subsection{Experiments Settings}

\subsubsection{Datasets}
\paragraph{Kumar} \cite{kumar2017dataset} contains 30 H\&E stained images (size: 1000$\times$1000) with 21,623 annotated nuclei. The training and test set contains 16 and 14 images, respectively.
\paragraph{CPM-17} \cite{vu2019methods} comprises 64 H\&E stained images (size: 500$\times$500 or 600$\times$600) with 7,570 annotated nuclei. Both the training and test sets contain 32 images.
\paragraph{PanNuke} \cite{gamper2019pannuke,gamper2020pannuke} is regarded as one of the most challenging datasets for simultaneous nucleus instance segmentation and classification. The dataset includes 7,899 H\&E stained images of size 256$\times$256 and 189,744 nuclei classified into five classes. We use the same three-fold cross-validation splits as \cite{gamper2020pannuke} and report the averaged results across these three splits.

\subsubsection{Evaluation Metrics}
Following prior works, we adopt Aggregated Jaccard Index (AJI) and Panoptic Quality (PQ) for comparison. Since AJI suffers from the over-penalization issue in overlapping regions \cite{graham2019hover}, we consider PQ as the primary metric.

\paragraph{Implementation details} is available in our supplementary material.

\begin{table}[t!]
	\centering
	\resizebox{0.98\linewidth}{!}{		
		\begin{tabular}{c|cc|cc}
			\toprule[1.5pt]
			\multirow{2}{*}{Method} & \multicolumn{2}{c|}{Kumar} &
			\multicolumn{2}{c}{CPM-17}\\
			\cline{2-5}
			& AJI & PQ & AJI & PQ \\
			\toprule[1.5pt]
			U-Net \cite{ronneberger2015u} & 0.556 & 0.478 & 0.666 & 0.625\\
			DCAN \cite{chen2016dcan} & 0.525 & 0.492 & 0.561 & 0.545\\
			Mask-RCNN \cite{he2017mask} & 0.546 & 0.509 & 0.684 & 0.674 \\
			DIST \cite{naylor2018segmentation} & 0.559 & 0.443 & 0.616 & 0.504 \\
			Micro-Net \cite{raza2019micro} & 0.560 & 0.519 & 0.668 & 0.661 \\
			CIA-Net \cite{zhou2019cia} & 0.620 & 0.577 & - & - \\
			Full-Net \cite{qu2019improving} & 0.601 & \underline{0.620} & 0.702 & 0.686 \\								
			Hover-Net \cite{graham2019hover} & 0.618 & 0.597 & 0.705 & 0.697 \\
			Triple U-Net \cite{zhao2020triple} & \underline{0.621} & 0.601 & 0.711 & 0.685 \\
			FEEDNet \cite{deshmukh2022feednet} & 0.616 & 0.613 & 0.701 & 0.705 \\
			HARU-Net \cite{chen2023enhancing} & 0.613 & 0.572 & \underline{0.721} & 0.701 \\
			PointNu-Net \cite{yao2023pointnu} & 0.606 & 0.603 & 0.712 & \underline{0.706} \\
			PromptNucSeg-B (Ours) & 0.614 & 0.620 & 0.731 & 0.726 \\
			\rowcolor{blue!10} PromptNucSeg-B + TTA & 0.620 & 0.627 & 0.734 & 0.730 \\ 
			PromptNucSeg-L (Ours) & 0.621 & 0.626 & 0.734 & 0.730 \\
			\rowcolor{blue!10} PromptNucSeg-L + TTA & 0.625 & 0.631 & 0.740 & 0.735 \\
			PromptNucSeg-H (Ours) & \textbf{0.622} & \textbf{0.627} & \textbf{0.740} & \textbf{0.733} \\
			\rowcolor{blue!10} PromptNucSeg-H + TTA & 0.624 & 0.631 & 0.743 & 0.737 \\
			\bottomrule[1.5pt]
	\end{tabular}}
	\caption{Performance comparison on Kumar and CPM-17 datasets. The experimental results rendered with \textcolor{blue!50}{blue} are omitted from the comparison because they are obtained with test-time augmentation (TTA), which is also employed by Full-Net \protect\cite{qu2019improving}. The optimal results are in \textbf{bold} while the previous best arts are \underline{underlined}.}
	\label{tab:kumar_cpm17}
	\vspace{-5pt}
\end{table}

\subsection{Comparison with SOTA Methods}
We use PromptNucSeg-B/L/H to differentiate our method with fine-tuned SAM-B/L/H as the nucleus segmentor. Tab.~\ref{tab:pannuke} shows the quantitative comparison results of our approach with SOTA methods on the challenging PanNuke dataset. Without additional techniques such as stain normalization, oversampling or adding an auxiliary tissue classification branch \cite{horst2023cellvit}, PromptNucSeg-H outperforms the previous best models by 1.1 bPQ and 1.4 mPQ. Moreover, we report the detection and segmentation performance of various methods for each type of nuclei in Tabs.~\ref{tab:pannuke_det} and \ref{tab:pannuke_seg}. In a nutshell, our method achieves the highest F1 scores across all five classes for nucleus detection and the highest PQ scores for four out of the five categories in terms of nucleus segmentation. Tab.~\ref{tab:kumar_cpm17} exhibits the comparison results on the Kumar and CPM-17 benchmarks. In case of the Kumar dataset, our method outshines the runner-up by 0.1 points on AJI and 0.6 points on PQ. Moreover, it demonstrates a substantial improvement on the CPM-17 dataset, exceeding the second-highest AJI and PQ scores by 1.9 and 2.8 points, respectively. Due to the limited pages, we present the qualitative comparison results in the supplementary material.

We further analyze the model size, computational cost and inference efficiency of different methods on the PanNuke dataset in Tab.~\ref{tab:efficiency}. The counterparts demonstrate significantly higher MACs since they generally adopt the U-Net \cite{ronneberger2015u} architecture with progressive upsampling to regress high-resolution nuclear proxy maps. Besides, they manifest slower inference speed due to the accompanying CPU-intensive post-processing steps. In comparison, PromptNucSeg is cost-effective and efficient since it predicts nuclei prompts and their associated masks directly from hidden features of low resolution.
\begin{table}[t]
	\centering
	\resizebox{0.99\linewidth}{!}{		
		\begin{tabular}{c|cccc}
			\toprule[1.5pt]
			Method & Params (M) & MACs (G) & FPS & mPQ \\
			\shline
			StarDist & 122.8 & 263.6 & 17 & 0.4744\\
			HoVer-Net & 37.6 & 150.0 & 7 & 0.4629 \\
			CPP-Net & 122.8 & 264.4 & 14 & 0.4847 \\
			PointNu-Net & 158.1 & 335.1 & 11 & 0.4957 \\
			CellViT-B & 142.9 & 232.0 & 20 & 0.4923 \\
			PromptNucSeg-B & 145.6 & 59.0 & 27 & 0.5095 \\
			\bottomrule[1.5pt]
	\end{tabular}}
	\caption{Comparison of model size, computational cost, efficiency and performance on the PanNuke dataset. All metrics are measured on a single NVIDIA RTX 3090 GPU.}
	\vspace{-5pt}
	\label{tab:efficiency}
\end{table}

\subsection{Ablation Studies}
\begin{figure}[t!]
	\centering		
	\includegraphics[width=0.9\linewidth]{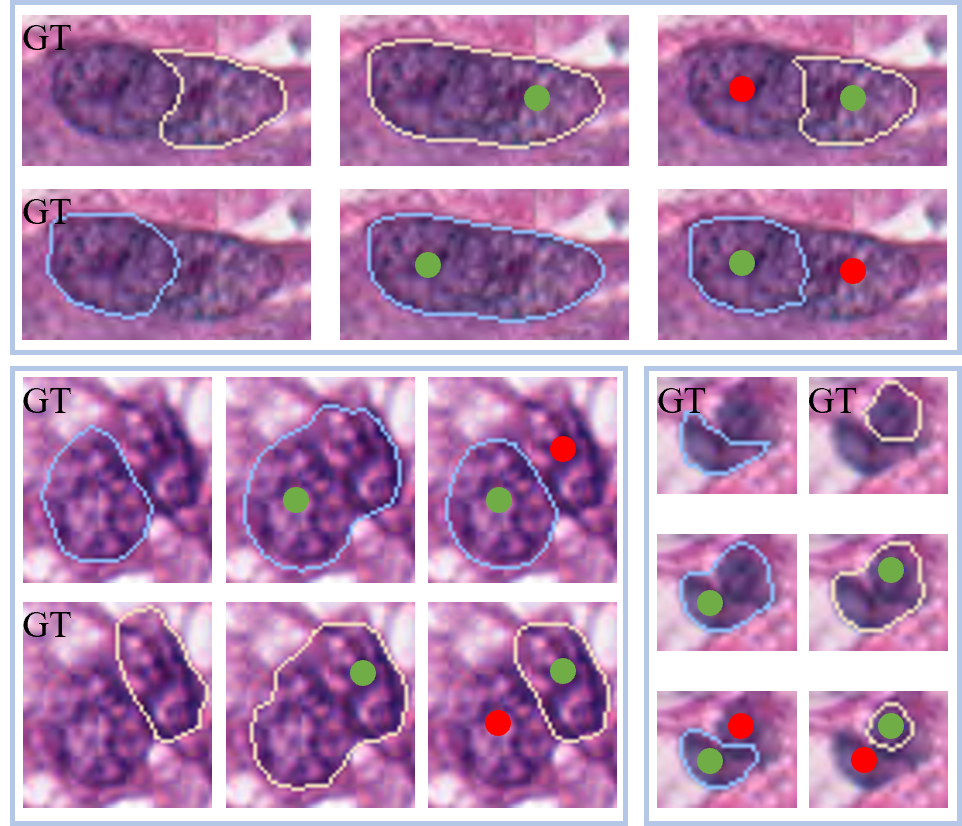}
	\vspace{-5pt}
	\caption{Three cases of our method w/ and w/o using negative prompts. The images marked by GT imply the ground-truth nuclear boundaries, while the others indicate predicted outcomes given different types of prompts. \textcolor{green}{$\mdlgblkcircle$} Positive prompt	\textcolor{red}{$\mdlgblkcircle$} Negative prompt}
	\label{fig:vis_np}
	\vspace{-5pt}
\end{figure}

On top of PromptNucSeg-H, we ablate the effect of our proposed modules on the CPM-17 dataset, which involve fine-tuning SAM (FT), auxiliary task learning of nuclear region segmentation (AUX), mask-aided prompt filtering (MAPF), and incorporation of negative prompts (NP). The experimental results in Tab.~\ref{tab:ablation} demonstrate that all the proposed modules contribute to improving the performance of our model. Fig.~\ref{fig:vis_np} visualizes the effect of NP.
\begin{table}[t!]
	\centering
	\small{
		\begin{tabular}{cccc|cc}
			\toprule[1.5pt]
			FT & AUX & MAPF & NP & AJI & PQ \\
			\shline
			& & & & 0.319 & 0.223 \\		
			$\checkmark$ & & & & 0.728 & 0.723 \\
			$\checkmark$ & $\checkmark$ & & & 0.734 & 0.727\\
			$\checkmark$ & $\checkmark$ & $\checkmark$ & & 0.737 & 0.731 \\
			$\checkmark$ & $\checkmark$ & $\checkmark$ & $\checkmark$ & 0.740 & 0.733 \\
			\bottomrule[1.5pt]			
		\end{tabular}
	}
	\caption{Effect of our proposed modules.}
	\vspace{-10pt}
	\label{tab:ablation}
\end{table}

\paragraph{Effect of the number of negative prompts}
We investigate the performance of PromptNucSeg-H with varying numbers of negative prompts on the CPM-17 dataset, as detailed in Tab.~\ref{tab:ablation_k}.

We initially assess the practical performance of our method by feeding predicted nuclei prompts into the segmentor. The results in Rows 1-3 discover that adding negative prompts solely in the inference stage cannot enhance the model's performance. We speculate that fine-tuning with only positive prompts results in a catastrophic forgetting about the effect of negative prompts. Comparing Rows 5 and 6, as well as Rows 8 and 9, we find that employing 1 negative prompt yields better outcomes than using 2 negative prompts. We posit that this discrepancy arises from the inherent noise in predicted prompts, the introduction of which is particularly notable when using two negative prompts.

To verify our suspicions, we further test the "oracle" performance of our method by using ground-truth nuclear centroids as prompts for the segmentor. Comparing Rows 2 and 5, we observe the "oracle" performance significantly improves when negative prompts are integrated into the fine-tuning process. This observation confirms the existence of the catastrophic forgetting problem explained earlier. Examining Rows 4-6 and 7-9, we find that when the prompts are noise-free, the "oracle" performance continually improves with the number of negative prompts. This finding substantiates our second suspicion. 

The substantial gaps between practical and "oracle" performance underscore the impact of prompt quality on the overall system performance. Given that training the prompter necessitates only nuclei point annotations, it is promising to improve the nucleus instance segmentation outcomes in a cost-effective scheme by bolstering the prompter's accuracy with more budget-friendly point labels.

\paragraph{Which module should undertake the nucleus classification task? Answer is the prompter.} In prior experiments, we employ the prompter for nucleus classification. Here we explore the performance of PromptNucSeg when training the prompter in a class-agnostic manner and transferring the classification function to the segmentor. To adapt the class-agnostic SAM for nucleus classification, we append a [cls] token to the mask decoder and update it in the same way as the [mask] and [IoU] tokens. Subsequently, the updated [cls] token is fed into a MLP head to predict the categorical logits. We incorporate a multi-class focal loss of weight 1 into Eq.~\ref{eq:loss} to supervise the classification learning. Tab.~\ref{tab:cls} displays the performance of PromptNucSeg-H on the PanNuke dataset when the prompter and segmentor are responsible for nucleus classification, respectively. The results suggest a slight performance advantage of using the prompter for nucleus classification over the segmentor.

\paragraph{Further ablation studies} 
By sharing the image encoder of the prompter and segmentor, we explore the performance of PromptNucSeg trained in an end-to-end style. In addition, we scrutinize the impact of overlapping size between sliding windows on model performance. The results and discussions are detailed in the supplementary material.
\begin{table}[t!]
	\centering
	\resizebox{0.8\linewidth}{!}{		
		\begin{tabular}{c|cc|cc|cc}
			\toprule[1.5pt]
			\multirow{3}{*}{Row id} & \multicolumn{2}{c|}{\# NP} &  \multicolumn{4}{c}{Source of Prompts} \\
			\cline{2-7}
			& \multirow{2}{*}{Train} & \multirow{2}{*}{Test} & \multicolumn{2}{c|}{Pred} & \multicolumn{2}{c}{GT} \\
			\cline{4-7}
			& & & AJI & PQ & AJI & PQ \\
			\shline
			1 & 0 & 0 & 0.737 & 0.731 & 0.779 & 0.772 \\
			2 & 0 & 1 & 0.737 & 0.730 & 0.794 & 0.782 \\
			3 & 0 & 2 & 0.735 & 0.729 & 0.793 & 0.778 \\
			\shline
			4 & 1 & 0 & 0.737 & 0.733 & 0.780 & 0.772 \\
			5 & 1 & 1 & 0.740 & 0.733 & 0.804 & 0.790 \\
			6 & 1 & 2 & 0.736 & 0.729 & 0.808 & 0.791 \\
			\shline
			7 & 2 & 0 & 0.739 & 0.731 & 0.778 & 0.769 \\
			8 & 2 & 1 & 0.740 & 0.732 & 0.804 & 0.790 \\
			9 & 2 & 2 & 0.738 & 0.729 & 0.811 & 0.796 \\
			\bottomrule[1.5pt]
	\end{tabular}}
	\vspace{-5pt}
	\caption{Effect of the number of negative prompts.}
	\vspace{-5pt}
	\label{tab:ablation_k}
\end{table}
\begin{table}[t!]
	\centering
	\resizebox{1.0\linewidth}{!}{		
		\begin{tabular}{c|cc|ccccc}
			\toprule[1.5pt]
			\multirow{2}{*}{Classifier} & \multicolumn{2}{c|}{Tissue} & \multicolumn{5}{c}{Nucleus} \\
			\cline{2-8}
			& bPQ & mPQ & Neop. & Epit. & Infl. & Conn. & Dead \\
			\shline
			Prompter & 0.692 & 0.512 & 0.598 & 0.582 & 0.441 & 0.433 & 0.161 \\									
			Segmentor & 0.688 & 0.506 & 0.587 & 0.587 & 0.423 & 0.431 & 0.157 \\
			\bottomrule[1.5pt]	
	\end{tabular}}
	\vspace{-5pt}
	\caption{Model performance with the prompter and segmentor as nucleus classifier, respectively.}
	\vspace{-5pt}
	\label{tab:cls}
\end{table}

\section{Conclusion}
In this paper, we have presented PromptNucSeg, a SAM-inspired method for automatic nucleus instance segmentation in histology images. Architecturally, PromptNucSeg consists of two parts: a prompter generating a distinct point prompt for each nucleus, and a segmentor predicting nuclear masks driven by these prompts. Extensive experiments across three benchmarks document the superiority of PromptNucSeg.
\paragraph{Limitations and future work.} Despite the compelling accuracy and speed, PromptNucSeg exhibits a slight drawback in terms of model size compared to its counterparts, resulting in increased storage and transmission costs. For future work, we plan to diminish the model size through pruning, devise more powerful prompter, and explore the performance of PromptNucSeg built up other SAM-like pre-trained models, such as Semantic-SAM \cite{li2023semantic}, MedSAM \cite{ma2023segment} and SAM-Med2D \cite{cheng2023sam}.

\bibliographystyle{named}
\bibliography{ijcai24}


\end{document}